\documentclass[lettersize,journal]{IEEEtran}
\usepackage{amsmath,amsfonts}
\usepackage{algorithmic}
\usepackage{algorithm}
\usepackage{array}
\usepackage[caption=false,font=normalsize,labelfont=sf,textfont=sf]{subfig}
\usepackage{textcomp}
\usepackage{stfloats}
\usepackage{url}
\usepackage{verbatim}
\usepackage{graphicx}
\usepackage{cite}

\usepackage{amssymb}
\usepackage{multirow}
\usepackage{booktabs}
\usepackage{ragged2e}
\usepackage{subfloat}
\usepackage{subfig}

\usepackage{fancyhdr}

\begin{document}

\title{Adversarial Representation with Intra-Modal and Inter-Modal Graph Contrastive Learning for Multimodal Emotion Recognition}

\author{Yuntao~Shou, Tao~Meng,
	Wei~Ai,
	Nan Yin, and~Keqin~Li,~\IEEEmembership{Fellow,~IEEE}
	\thanks{Corresponding Author: Tao Meng~(mengtao@hnan.edu.cn)}
		\IEEEcompsocitemizethanks{\IEEEcompsocthanksitem Y. Shou,~T. Meng, and~W. Ai are with School of computer and Information Engineering, Central South University of Forestry and Technology, Hunan 410004,
		China. (mengtao@hnan.edu.cn,\ yuntaoshou@csuft.edu.cn,~aiwei@hnan.edu.cn)
		\IEEEcompsocthanksitem N. Yin is with Mohamed bin Zayed University of Artificial Intelligence, UAE. (nan.yin@mbzuai.ac.ae)
		\IEEEcompsocthanksitem K. L is with the Department of Computer Science, State University of New York, New Paltz, New York 12561, USA. (lik@newpaltz.edu)}}

\markboth{Please submit the 
	manuscript to the Special Issue on Graph Learning}%
{Shell \MakeLowercase{\textit{et al.}}: A Sample Article Using IEEEtran.cls for IEEE Journals}


\maketitle

\begin{abstract}
With the release of increasing open-source emotion recognition datasets on social media platforms (e.g., Weibo, Twitter, and Meta, etc.) and the rapid development of computing resources, multimodal emotion recognition tasks (MER) have begun to receive widespread research attention. The MER task extracts and fuses complementary semantic information from different modalities, which can classify the speaker's emotions. However, the existing feature fusion methods have usually mapped the features of different modalities into the same feature space for information fusion, which can not eliminate the heterogeneity between different modalities. Therefore, it is challenging to make the subsequent emotion class boundary learning. To tackle the above problems, we have proposed a novel Adversarial Representation with Intra-Modal and Inter-Modal Graph Contrastive for Multimodal Emotion Recognition (AR-IIGCN) method. Firstly, we input video, audio, and text features into a multi-layer perceptron (MLP) to map them into separate feature spaces. Secondly, we build a generator and a discriminator for the three modal features through adversarial representation, which can achieve information interaction between modalities and eliminate heterogeneity among modalities. Thirdly, we introduce contrastive graph representation learning to capture intra-modal and inter-modal complementary semantic information and learn intra-class and inter-class boundary information of emotion categories. Specifically, we construct a graph structure for three modal features and perform contrastive representation learning on nodes with different emotions in the same modality and the same emotion in different modalities, which can improve the feature representation ability of nodes. Finally, we use MLP to complete the emotional classification of the speaker. Extensive experimental works show that the ARL-IIGCN method can significantly improve emotion recognition accuracy on IEMOCAP and MELD datasets. Furthermore, since AR-IIGCN is a general multimodal fusion and contrastive learning method, it can be applied to other multimodal tasks in a plug-and-play manner, e.g., humour detection. 
\end{abstract}


\section{Introduction}
\IEEEPARstart{T}he multimodal emotion recognition task (MER) can combine the semantic information with different modal features (e.g., text, video, audio, etc.), which identifies the emotion of the speaker at the current moment \cite{9358000}, \cite{ai2023gcn}, \cite{meng2023deep}, \cite{shou2023comprehensive}. With the continuous development of deep learning technology and computing resources, MER has been increasingly used in many practical social media scenarios. For example, in a human-computer dialogue system, the interactive system can obtain the user's current emotional state of the user according to the data analysis of the human-computer dialogue. Then, it can generate words to fit the scene. Therefore, accurate identification of the user's emotional state has become a practical application value \cite{khare2020time}.

\begin{figure}
	\centering
	\includegraphics[width=1\linewidth]{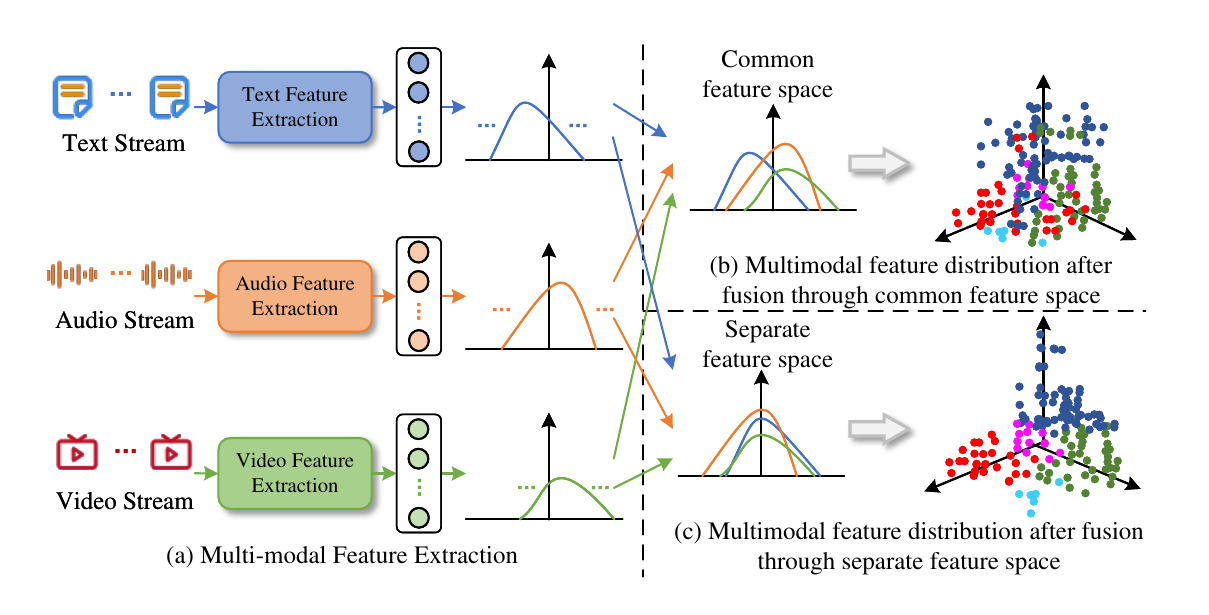}
	\caption{Illustrative example of the effect of different feature fusion methods on sentiment classification. (a) The feature extraction process for text, video and audio modalities. (b) The model learns emotion class boundaries in a feature fusion manner to map a common feature space. (c) The model learns emotion class boundaries in a feature fusion manner to map a separate feature space. Specifically, there is heterogeneity among modalities in feature fusion with common space, which leads to the misalignment of peaks between modalities.} 
	\label{fig:}
\end{figure}

However, MER can eliminate the multimodal heterogeneous data modality gap because video, audio, and text feature distributions are inconsistent in space. The current mainstream feature fusion method can eliminate the gap of different modal data, which is to map into the same feature space for feature representation \cite{zhao2022cauain}. For example, Tensor Fusion Network (TFN) \cite{zadeh2017tensor} uses the tensor outer product operation to map different modal features into a three-dimensional feature space, which is used for the fusion representation of multimodal feature vectors. Low-rank Fusion Network (LFN) \cite{liu2018efficient} utilizes low-rank decomposition operations to combine correlated feature vectors highly and fuse three modal features. However, the above methods forcibly map different modal features into a common representation space, which can not eliminate their heterogeneity. It is presented the Fig. 1 as an example. Fig. 1(b) shows the distribution of multimodal features mapped to a unified space. Fig. 1(c) illustrates the distribution of multimodal features obtained through adversarial representation learning. The above phenomenon is indicated that the spatial distribution learned by adversarial learning is much discriminative \cite{shou2023czl}, \cite{shou2023graph}.

For the problem with existing deep learning methods, they have failed to capture intra-class and inter-class semantic information, which is differentiated \cite{meng2023multi}, \cite{ying2021prediction}, \cite{shou2022object}. However,  current mainstream research has mainly focused on capturing complementary semantic information between modalities, which ignores the relationship between modalities and emotion categories \cite{shou2022conversational}. For instance, Hu et al. \cite{hu2021mmgcn} proposed Multimodal Fusion via Deep Graph Convolution Network (MMGCN) to fuse dialogue relations and complementary semantic information of different modalities through GCN. Moreover, Liu et al. \cite{liu2023multi} proposed a Multimodal Fusion Network (MFN) and used an attention mechanism to consider the importance of different modalities. It was obtained as a multimodal fusion vector with modal interactions, which was challenging to learn clear class boundaries between different emotion categories. However, many studies \cite{kosti2017emotion}, \cite{zhang2020emotion}, \cite{lee2019context} have investigated capturing the relationship between different modalities and emotion categories, which improves emotion classification.

To eliminate the heterogeneity between different modalities and capture the intra-modal, inter-modal complementary semantic information, intra-class, and inter-class differences, it is still a problem to be solved.

To solve the above problem, we propose a novel Adversarial Representation with Intra-Modal and Inter-Modal Graph Contrastive Learning for Multimodal Emotion Recognition, i.e., AR-IIGCN. Firstly, we use RoBERTa, DenseNet, and Bi-LSTM-based Encoder to obtain semantic information in text, video, and audio. Secondly, we input the extracted three modality features into a multi-layer perceptron (MLP) to map them into separate feature spaces. Thirdly, we build a generator and discriminator for the three modal features. It is used adversarial learning to achieve cross-modal feature fusion and eliminate the heterogeneity between different modalities. Fourthly, we construct a new graph contrastive representation learning architecture, which captures complementary semantic information within and between modalities and intra-class and inter-class differences. It is performed contrastive representation learning on nodes with different emotions in the same modality and nodes with the same emotion in different modalities, which obtains a more precise representation of the boundary distribution. Finally, we have successfully used MLP for emotion classification.

\subsection{Our Contributions}
Therefore, MER can consider eliminating the heterogeneity among the three modalities of video, audio, and text, which can capture the complementary semantic information within and between modalities and the intra-class and inter-class differences. Inspired by the above analysis, we propose a novel Adversarial Representation with Intra-Modal and Inter-Modal Graph Contrastive Learning for Multimodal Emotion Recognition (AR-IIGCN), which learns better emotion class boundary information. The main contributions of this paper are summarized as follows:

\begin{itemize}
	\item A novel Adversarial Representation Learning with Intra-Modal and Inter-Modal Graph Contrastive Learning for Multimodal Emotion Recognition architecture is investigated, i.e., ARL-IIGCN. ARL-IIGCN can learn better emotion class boundary information.
	
	\item A new cross-modal feature fusion method with adversarial learning is designed to capture complementary semantic information between modalities, which can eliminate heterogeneity among modalities.
	
	\item A novel graph contrastive representation learning framework is investigated to enhance the correlation of intra-modal and inter-modal semantic information, which can learn the intra-class and inter-class differences.
	
	\item A new multiple loss is designed to graph contrastive representation learning. It is forced positive embeddings and closed to anchor embeddings, and negative embeddings are farther away from anchor embeddings.
	
	\item Finally, extensive experimental works are conducted on two benchmark datasets, i.e., MELD and IEMOCAP. The experimental results show that the emotion recognition effect of ARL-IIGCN is better than the existing comparison algorithms.
	
\end{itemize}

The rest of this paper is organized as follows. Section 2 presents the related work of prior MER. Section 3 describes the multi-modal emotion recognition task and presents the multi-modal data processing flow. Section 4 illustrates the proposed neural network AR-IIGCN. Section 5 describes the datasets and evaluation metrics used. The related experimental results and discussion on the IEMOCAP and MELD datasets are shown in Section 6.  Finally, we conclude our work and illustrate future work.

\section{Related work}

\subsection{Multimodal Emotion Recognition in Conversation}
As an interdisciplinary study (e.g., brain science and cognitive science, etc.), MER has received extensive attention from researchers \cite{9839579}. The current mainstream MER research has mainly included sequential context modelling, speaker relationship modelling, and multimodal feature fusion modelling. The sequential context modelling method has mainly combined the semantic information of the context, which classifies the emotion at the current moment. The speaker relationship modelling method extracts the semantic information of the dialogue relationship between speakers through the graph convolution operation. The multimodal feature fusion modelling method achieves cross-modal feature fusion by capturing intra-modal and inter-modal complementary semantic information.

In the modelling method based on sequential context, Poria et al. \cite{poria-etal-2017-context} proposed bidirectional long short-term memory (Bi-LSTM), which can extract contextual semantic information of forward and reverse sequence. However, bc-LSTM has a limited ability to model long-distance context dependencies. To respond to the above problems, Beard et al. \cite{beard-etal-2018-multi} proposed recursive multi-attention (RM), which is used multi-gated memory units to update the memory network iteratively. Therefore, it is realized in the memory of global context information. Although sequential context-based modelling can achieve specific results in emotion recognition, it ignores the intra-modal and inter-modal complementary semantic information.

In the modelling method based on multimodal feature fusion, Zadeh et al. \cite{zadeh2017tensor} proposed Tensor Fusion Network (TFN) to map multimodal features into three-dimensional space through tensor outer product operation, which realizes information interaction between multimodal features. However, the feature dimension of TFN is high, which is prone to an overfitting effect. To alleviate the problems of TFN, Liu et al. \cite{liu2018efficient} proposed a Low-rank Fusion Network (LFN) to realize dimensionality reduction of tensors through low-rank decomposition operations, which has achieved performance improvement in emotion recognition. Moreover, Hu et al. \cite{hu2021mmgcn} proposed Multimodal Fusion via Deep Graph Convolution Network (MMGCN), which can effectively utilize the complementary semantic information between multimodal features. Although the above methods can achieve cross-modal feature fusion, they have mapped the features of different modalities into the same feature space. It is challenging to eliminate the heterogeneity between different modalities.

In the modelling method based on speaker relationship, Ren et al. \cite{9556142} proposed Latent Relation-Aware Graph Convolutional Network (LR-GCN) and constructed a speaker relation graph. Then, a multi-head attention mechanism was introduced to capture latent relations between utterances. However, the fully connected graphs introduce noise information. Nie et al. \cite{nie2020c} proposed that the Correlation-based Graph Convolutional Network (C-GCN) method captured the correlation inter and intra-modalities, which realized the effective use of multimodal information. Although the modelling method based on speaker relationships can fully use the semantic information of speaker dialogue relationships and cross-modal semantic information, it ignores the differences between different emotion categories.

\subsection{Generative Adversarial Learning}
In multimodal emotion recognition, data imbalance is a common problem, which leads to biased learning of the model \cite{9868807}. Therefore, researchers have begun adopting generative adversarial learning to generate new samples that fit the original data distribution. Specifically, previous work generates new samples by minimizing the data distribution learned by the generator and the discriminator.

Su et al. \cite{9815553} proposed Corpus-Aware Emotional CycleGAN (CAEmoCyGAN) and introduced a target-to-source generator to generate new samples that more closely matched the original data distribution. CAEmoCyGAN can enhance the ability of the model to learn unbiased representations. Moreover, Chang et al. \cite{9609567} proposed Adversarial Cross Corpora Integration (ACCI) and used an adversarial autoencoder to generate samples with contextual semantic information. It used emotion labels as additional constraints for the model. Although new samples are generated by generative adversarial learning, it can effectively alleviate the data imbalance problem, which can not eliminate the heterogeneity among data of different modalities.

\subsection{Contrastive Learning}
Self-supervised learning (SL) is an essential branch of deep learning (DL), which has received increasing research attention because of its powerful ability to learn representations. Contrastive representation learning (CRL) is one of the representative methods. Specifically, CRL learns discriminative features by continuously shrinking the distance (e.g., Euclidean distance and Mahalanobis distance, etc.) between positive and negative samples and expanding the distance between positive and negative samples. Previous works have obtained representations of features by maximizing the mutual information (MI) between model inputs and learned representations.

Li et al. \cite{li2021contrastive} proposed contrastive predictive coding (CPC) to address the lack of large-scale datasets for emotion recognition tasks. Through unsupervised contrastive representation learning, CPC can learn latent emotional semantic information from unlabeled data. Furthermore, Kim et al. \cite{kim2021contrastive} proposed contrastive adversarial learning (CAL) to solve the problem of existing methods, which relied on supervised information. CAL is used to learn complex semantic emotional information by comparing samples with strong emotional features and samples with weaker emotions. Wang et al. \cite{9887975} designed a new architecture composed of three networks (i.e., FacesNet, SceneNet, and ObjectsNet) to improve the feature fusion ability of the model. It was used to solve the problem of missing essential semantic information. Although contrastive representation learning can enhance the representation of emotional information, the above methods ignore the intra-modal and inter-modal information interaction and intra-class and inter-class contrastive representation learning.

\section{Preliminary Information}
This section defines the Multimodal Conversational Emotion Recognition Task (MCER) in mathematical terms. In addition, It is described the data preprocessing methods of different modalities as follows: (1) Word Embedding: To eliminate the ambiguity of words, this paper is used RoBERTa \cite{liu2019roberta} to obtain the embedding representation of word vectors (2) Visual Feature Extraction: It is used DenseNet \cite{huang2017densely} to capture deeper image features in videos and reduce the introduction of noisy information. (3) Audio Feature Extraction: The encoder architecture is adopted \cite{NEURIPS2018_6832a7b2} to extract audio signals from different speakers.

\subsection{Problem Definition}
The task of MER is to predict the sentiment class of test utterances in the dataset. It is assumed that there are $N$ speakers in a dialogue context, and the set of speakers can be represented as $S=\{P_1,P_2,\ldots,P_N\}$. The dialogue context is sorted according to the order in which each speaker speaks, which can be expressed as $D=\{d_1,d_2,\ldots,d_T\}$, where $T$ is the total number of test utterances. This paper defines a mapping function $f$ to construct the index relationship between speakers and test utterances, which is expressed as $H=\{\delta_{f_1 },\delta_{f_2 },\ldots,\delta_{f_i }\}$, where $i$ represents the $i$-th test utterances. Our task is to classify the emotion of each utterance $\delta_{f_i }$.

\subsection{Multimodal Feature Extraction}
The experimental datasets of IEMOCAP and MELD can consist of three modalities, which are stored in text, video, and audio, respectively. For the features of different modalities, a specific data preprocessing method for feature extraction is used to obtain feature vector representations with less noise and rich semantic information. It is described the features encoded for each modality as follows.

\subsubsection{Word Embedding}
To disambiguate words and obtain feature vectors with rich semantic information, we have used the RoBERTa model \cite{liu2019roberta} to encode words. In this paper, it is used sentence-level encoding to encode each utterance of the speaker, and obtain a contextual semantic representation $\varphi_i=\{\varphi_i^1,\varphi_i^n,…,\varphi_i^m\}$ contains the entire sentence. Where $m$ is the dimension of word embedding. Due to limited computing resources, it is collected the first 100-dimensional vectors encoded by the RoBERTa model as our word embedding representation $\xi_u$.

\subsubsection{Visual Feature Extraction}
The speaker's facial expression and behaviour reflect the inner emotional state of the speaker. Therefore, we capture the speaker's facial expressions and action changes from the video frames, thereby extracting semantic information related to the speaker's emotional changes. Moreover, it is used the DenseNet model to obtain a 512-dimensional feature vector $\xi_v$.

\subsubsection{Audio Feature Extraction}
The fluctuation of the voice in the audio signal also reflects the emotional changes in the speaker's heart. Sometimes, a person's actions may not truly reflect his emotions, but the tone changes cannot be faked. Therefore, we have used the encoder structure to extract the speaker's audio features $\xi_a$, where $\xi_a$ is a 100-dimensional feature vector. Specifically, the encoder comprises Bi-LSTM, an attention layer, and a fully connected layer.

\section{Methodology}
To increase the performance of multimodal emotion recognition, we have proposed a novel Adversarial Representation with Intra-Modal and Inter-Modal Graph Contrastive Learning for Multimodal Emotion Recognition, namely AR-IIGCN. The overall architecture of AR-IIGCN is shown in Fig. 1. The AR-IIGCN consists of data preprocessing, multimodal feature fusion, graph contrastive representation learning and emotion classification. In the data preprocessing stage, RoBERTa, DenseNet and Encoder are used to extract text, video and audio features. In the multimodal feature fusion stage, we build a generator and a discriminator for text, video, and audio to remove modality heterogeneity in an adversarial learning manner. In the graph contrastive learning stage, we have constructed two contrastive losses to perform intra-modal and inter-modal contrastive learning and intra-class and inter-class contrastive learning, respectively. Finally, we used MLP for emotion classification in the emotion classification stage.

\subsection{The Design of the AR-IIGCN Structure}
\begin{figure*}
	\centering
	\includegraphics[width=1.015\linewidth]{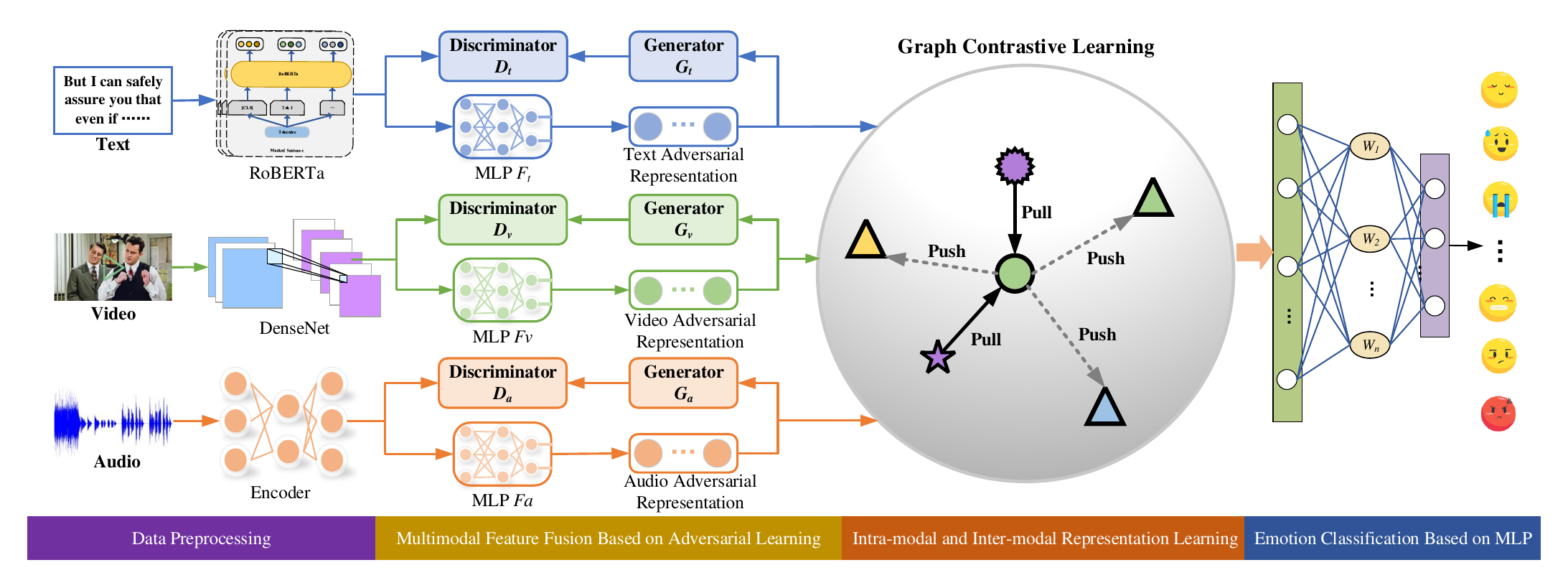}
	\caption{The overall framework of the Adversarial Representation Learning with Intra-Modal and Inter-Modal Graph Contrastive Learning consists of a data preprocessing layer, a multimodal feature fusion layer, a graph contrastive representation learning layer, and an emotion classification layer.}
	\label{fig:erc}
\end{figure*}

\subsubsection{TGAN: Tri-modal Generative Adversarial Networks}
Firstly, the MLP is used to dimensionally align the three modality features and map them into three separate feature spaces. The formulas are as follows:
\begin{equation}
	\begin{aligned}
		& \tilde{\xi}_u=f_{m l p}\left(\xi_u\right) \in \mathbb{R}^d \\
		& \tilde{\xi}_v=f_{m l p}\left(\xi_v\right) \in \mathbb{R}^d \\
		& \tilde{\xi}_a=f_{m l p}\left(\xi_a\right) \in \mathbb{R}^d
	\end{aligned}
\end{equation}
where $d$ denotes the dimension that maps the three modal features to a separate representation space. $f_{mlp} (\cdot)$ represents the MLP layer.

Secondly, we build a text generator and a text discriminator. The input of the text generator is audio features $\tilde{\xi}_a$ and video features $\tilde{\xi}_v$. The input of the text discriminator is the fused features generated by the text generator containing three modal information. The objective optimization function of the text generator is shown as follows:
\begin{equation}
	\begin{aligned}
		&\min _{G_{T}} \mathcal{L}_{G e n}\left(G_T, D_T\right) \\ & =\mathbb{E}_{\tilde{\xi}_a \sim P_{\text {data }}\left(\tilde{\xi}_a\right)}\left[\log \left(1-D_T\left(G_T\left(\tilde{\xi}_a\right)\right)\right)\right] \\
		&+  \mathbb{E}_{\tilde{\xi}_v \sim P_{\text {data }}\left(\tilde{\xi}_v\right)}\left[\log \left(1-D_T\left(G_T\left(\tilde{\xi}_v\right)\right)\right)\right] \\
	\end{aligned}
\end{equation}
where $G_T$ and $D_T$ represent text generator and text discriminator, $\tilde{\xi}_a \sim P_{\text {data }}$ represents sampling samples from the data that conforms to the audio feature distribution law, and $\tilde{\xi}_v \sim P_{\text {data }}$ represents sampling samples from the data that conforms to the video feature distribution law.

The objective optimization function of the text discriminator is discussed as follows:

\begin{equation}
	\begin{aligned}
		&\max _{D_{T}} \mathcal{L}_{ {Dis }}\left(G_T, D_T\right) \\ & =\mathbb{E}_{T \sim P_{\text {data }}(T)}\left[\log D_T(T)\right]\\ &+\mathbb{E}_{\tilde{\xi}_a \sim P_{\text {data }}\left(\tilde{\xi}_a\right)}\left[\log \left(1-D_T\left(G_T\left(\tilde{\xi}_a\right)\right)\right)\right] \\
		&	+ \mathbb{E}_{\tilde{\xi}_v \sim P_{\text {data }}\left(\tilde{\xi}_v\right)}\left[\log \left(1-D_T\left(G_T\left(\tilde{\xi}_v\right)\right)\right)\right]
	\end{aligned}
\end{equation}
where $\mathbb{E}_{T \sim P_{\text {data }}(T)}$ represents sampling samples from the data that conforms to the original data distribution.

Thirdly, it is build an audio generator and an audio discriminator. The input to the audio generator is text features and video features. The input of the audio discriminator is the fused features generated by the audio generator, which contains three modal information. The objective optimization function for the audio generator is calculated as follows:
\begin{equation}
	\begin{aligned}
		&\min_{G_{A}} \mathcal{L}_{ Gen }\left(G_A, D_A\right) \\ & =\mathbb{E}_{\tilde{\xi}_v \sim P_{\text {data }}\left(\tilde{\xi}_v\right)}\left[\log \left(1-D_A\left(G_A\left(\tilde{\xi}_v\right)\right)\right)\right] \\
		&	+  \mathbb{E}_{\tilde{\xi}_u \sim P_{\text {data }}\left(\tilde{\xi}_u\right)}\left[\log \left(1-D_A\left(G_A\left(\tilde{\xi}_u\right)\right)\right)\right] \\
	\end{aligned}
\end{equation}
where $G_A$ and $D_A$ represent audio generator and audio discriminator, $\tilde{\xi}_u \sim P_{\text {data }}$ represents sampling samples from the data that conforms to the text feature distribution law.

The objective optimization function for the audio discriminator is shown as follows:
\begin{equation}
	\begin{aligned}
	&\max _{D_{A}} \mathcal{L}_{D i s}\left(G_A, D_A\right) \\ & =\mathbb{E}_{A \sim P_{\text {data }}(A)}\left[\log D_A(A)\right]\\ &+\mathbb{E}_{\tilde{\xi}_v \sim P_{\text {data }}\left(\tilde{\xi}_v\right)}\left[\log \left(1-D_A\left(G_A\left(\tilde{\xi}_v\right)\right)\right)\right] \\
     &+  \mathbb{E}_{\tilde{\xi}_u \sim P_{\text {data }}\left(\tilde{\xi}_u\right)}\left[\log \left(1-D_A\left(G_A\left(\tilde{\xi}_u\right)\right)\right)\right]
     \end{aligned}
\end{equation}

Finally, we built a video generator and a video discriminator. The input of the video generator is text features and audio features. The input of the video discriminator is the fused features generated by the video generator containing three modal information. The objective optimization function for the video generator is as presented follows:
\begin{equation}
	\begin{aligned}
		& \min _{G_{V}} \mathcal{L}_{ {Gen }}\left(G_V, D_V\right) \\ & 
		=  \mathbb{E}_{\tilde{\xi}_a \sim P_{\text {data }}\left(\tilde{\xi}_a\right)}\left[\log \left(1-D_V\left(G_V\left(\tilde{\xi}_a\right)\right)\right)\right] \\
		&+  \mathbb{E}_{\tilde{\xi}_u \sim P_{\text {data }}\left(\tilde{\xi}_u\right)}\left[\log \left(1-D_V\left(G_V\left(\tilde{\xi}_u\right)\right)\right)\right] \\
	\end{aligned}
\end{equation}
where $G_V$ and $D_V$ represent video generator and video discriminator.

The objective optimization function for the video discriminator is shown as follows:
\begin{equation}
	\begin{aligned}
		&\max _{D_{V}} \mathcal{L}_{ {Dis }}\left(G_V, D_V\right) \\ & =\mathbb{E}_{V \sim P_{\text {data }}(V)}\left[\log D_V(V)\right]\\ &+\mathbb{E}_{\tilde{\xi}_a \sim P_{\text {data }}\left(\tilde{\xi}_a\right)}\left[\log \left(1-D_V\left(G_V\left(\tilde{\xi}_a\right)\right)\right)\right] 
\\ & +\mathbb{E}_{\tilde{\xi}_u \sim P_{\text {data }}\left(\tilde{\xi}_u\right)}\left[\log \left(1-D_V\left(G_V\left(\tilde{\xi}_u\right)\right)\right)\right]
     \end{aligned}
\end{equation}

It should be noted that after training the three-modal generative confrontation network, We input the output of MLP into GCL for training of subsequent tasks.

\subsubsection{Speaker Relation Graph Construction}
A graph structure is used to extract semantic information about speaker dialogue relations. Specifically, we construct a directed graph of speaker relations $\mathcal{G}_\mathcal{M}=\{\mathcal{V}_\mathcal{M},\mathcal{E}_\mathcal{M},\mathcal{R}_\mathcal{M},\mathcal{W}_\mathcal{M} \}$ for the three modal features of video, audio and text respectively, where $\mathcal{M} \in \{T,V,A\}$, the node $v_i^M (v_i^M \in V_\mathcal{M} )$ is composed of unimodal features $(i.e., \tilde{\xi}_a, \tilde{\xi}_v, \tilde{\xi}_u)$, the directed edge $r_{ij}^\mathcal{M} (r_{ij}^\mathcal{M} \in \mathcal{E}_\mathcal{M} )$ indicates that there is a dialogue relationship between the node $v_i^\mathcal{M}$ and the node $v_j^\mathcal{M}$, and $\omega_{ij}^\mathcal{M} (\omega_{ij}^\mathcal{M} \in \mathcal{W}_\mathcal{M},0 \leq \omega_{ij}^M \leq 1)$ is the weight of the edge $r_{ij} ^\mathcal{M}$, and $r^\mathcal{M} \in R_\mathcal{M}$ is the edge type. Since the computational complexity of GCN is $O(n^2)$, it leads to high computational resources required. Therefore, we set the context window size to 10.

To capture the key semantic information in the nodes, we use the attention mechanism is used to calculate the weight of the edge,, which  and performs  information aggregation according to the edge weight. Firstly, we use MLP to dynamically learn the correlation between node i and node j. The formula is defined as follows:
\begin{equation}
	\varepsilon_{i j}^{\mathcal{M}}=W_{\vartheta_1}^{\mathcal{M}}\left(\operatorname{GELU}\left(W_{\vartheta_2}^{\mathcal{M}}\left[\tilde{\xi}_i^{\mathcal{M}} \oplus \tilde{\xi}_j^{\mathcal{M}}\right]\right)\right)
\end{equation}
where $W_{\vartheta_1}^\mathcal{M}$, $W_{\vartheta_2}^\mathcal{M}$ are learnable network parameters, and $\oplus$ represents the vector concatenation operation.

Secondly, we use a softmax function to normalize the correlation between node $i$ and node $j$, which obtains the attention score for each edge. The formula is defined as follows:
\begin{equation}
	\omega_{i j}^\mathcal{M}=\operatorname{softmax}\left(\varepsilon_{i j}^{\mathcal{M}}\right)=\frac{\exp \left(\varepsilon_{i j}^{\mathcal{M}}\right)}{\sum_{j \in \mathcal{N}_i} \exp \left(\varepsilon_{i j}^{\mathcal{M}}\right)}
\end{equation}
where $\mathcal{N}_i$ represents the first-order neighbour nodes of node $i$. The larger $\omega_{ij}^\mathcal{M}$ represents the stronger correlation between node $i$ and node $j$.

Finally, we have updated the node representations using a GCN followed by a GELU activation function. The formula for GCN encoding is shown as follows:
\begin{equation}
	\begin{aligned}
		\psi_i^{\mathcal{M}}(t)&=G E L U\left(\sum_{r \in \mathcal{R}} \sum_{j \in \mathcal{N}_i^r} \frac{1}{\left|\mathcal{N}_i^r\right|}\left(\omega_{i j}^{\mathcal{M}} W_{\theta_1}^{\mathcal{M}} \psi_j^{\mathcal{M}}(t-1) \right.\right.	
		\\ & \left. +\omega_{i i}^{\mathcal{M}} W_{\theta_2}^{\mathcal{M}} \psi_i^{\mathcal{M}}(t-1)\right)\Bigg)
	\end{aligned}
\end{equation}
where $\mathcal{N}_i^r$ is the set of first-order neighbor nodes of node $i$ under the edge relationship $r\in R$, $|\mathcal{N}_i^r|$ is the modulus of $\mathcal{N}_i^r$, and $\psi_i^\mathcal{M} (t)$ is the feature vector encoded by GCN.

\begin{figure*}
	\centering
	\includegraphics[width=1\linewidth]{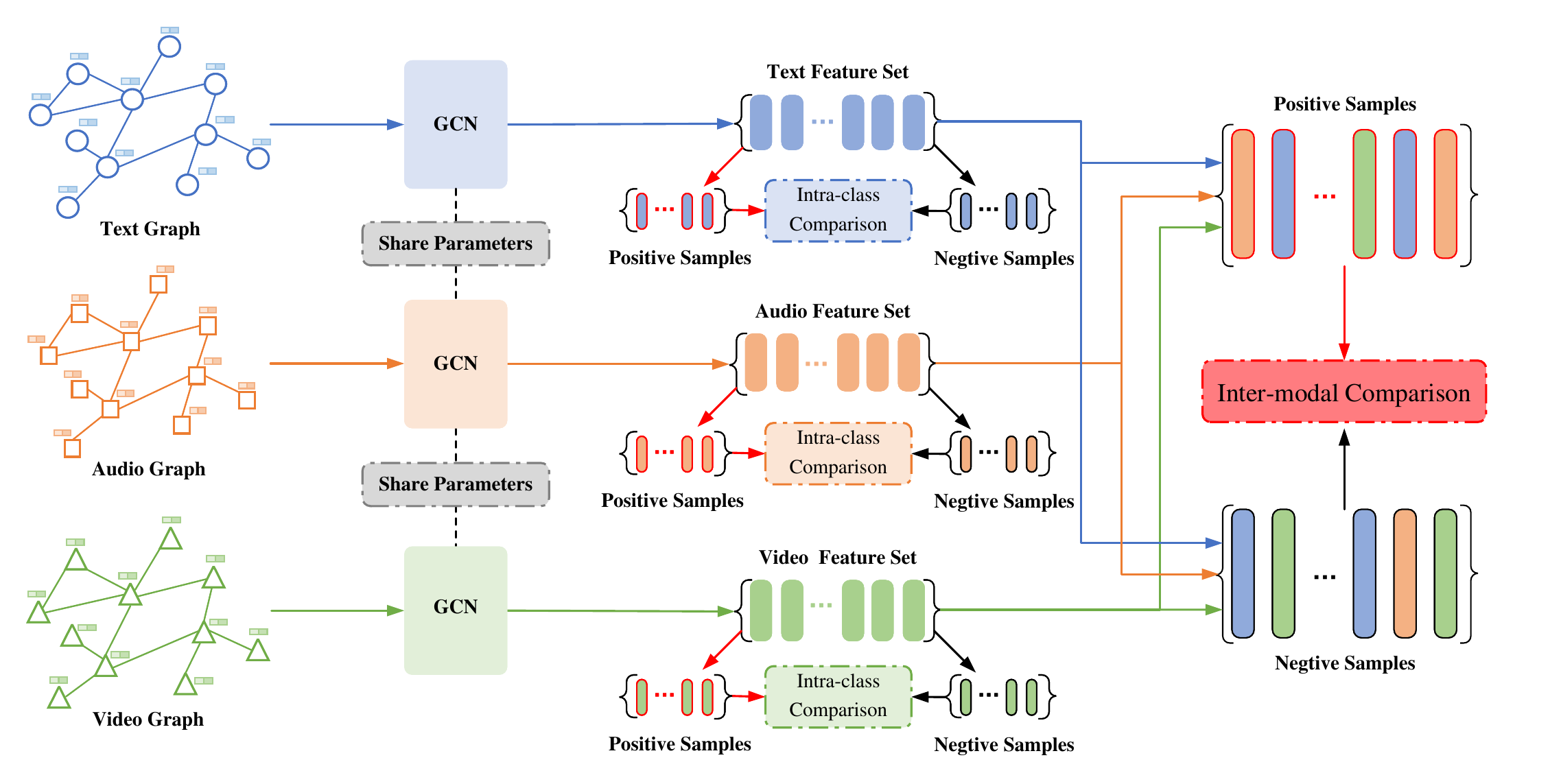}
	\caption{The overall graph contrastive representation learning process refers to intra-modal, inter-modal, and intra-class and inter-class comparisons.}
	\label{fig:}
\end{figure*}

\subsection{ICCL:  Intra-class and Inter-class Contrastive Learning}
ICCL aims to learn intra-class and inter-class semantic information with differences through contrastive learning. In IMCL, positive samples are represented by samples of the same modality and the same class, while samples of the same modality and different classes represent negative samples. The softmax function is used to normalize the representations of positive and negative samples so that the similarity between them ranges between 0 and 1. Specifically, the intra-class and inter-class contrastive loss are defined as follows:
\begin{equation}
	\mathcal{L}_{I C C L}=-\mathbb{E}_s\left[\frac{\sum_{i=1}^N\left(\mu^{\mathcal{M}}\right)^T \chi_i^{\mathcal{M}}}{\sum_{i=1}^N\left(\mu^{\mathcal{M}}\right)^T \chi_i^{\mathcal{M}}+\sum_{j=1}^M\left(\mu^{\mathcal{M}}\right)^T \delta_j^{\mathcal{M}}}\right]
\end{equation}
where $\chi_i^\mathcal{M}$ and $\delta_j^\mathcal{M}$ belong to samples of the same modality. 

However, if Eq. (11) is used as a contrastive loss, the model may fall into a local optimal solution. i.e., $(\mu^\mathcal{M} )^T \chi_i^\mathcal{M}$ can be minimized but $(\mu^\mathcal{M} )^T \delta_j$ cannot be maximized. The above situation occurs when the similarity between negative sample pairs is 0. No matter how much the similarity between positive sample pairs is, the contrastive loss of the model tends to the minimum value. The desired goal is that $(\mu^\mathcal{M})^T \chi_i^\mathcal{M}$ can be minimized and $(\mu^\mathcal{M} )^T \delta_j$ can be maximized. Therefore, we introduce a regularization term to ensure that the similarity between positive sample pairs can be maximized and the similarity between negative sample pairs can be minimized. The formula is defined as follows:

\begin{equation}
	\begin{aligned}
		\mathcal{L}_{I C C L}^R & =\mathbb{E}_S\left[\frac{1}{N} \sum_{i=1}^N\left\|\left(\mu^{\mathcal{M}}\right)^T \chi_i^{\mathcal{M}}-1\right\|^2\right] \\
		\tilde{\mathcal{L}}_{I C C L} & = \mathcal{L}_{I C C L}+\mathcal{L}_{I C C L}^R
	\end{aligned}
\end{equation}
where $\mathcal{L}_{I C C L}^R$ is the regularization loss for ICCL. With ICCL, the model can learn the intra-class and inter-class difference information.

\subsection{IMCL: Intra-modal and Inter-modal Contrastive Learning}
IMCL aims to learn complementary semantic information between modalities, which obtains a more discriminative embedding representation through a contrastive learning method. Specifically, in IMCL, positive samples are represented by samples of the same class in the same modality, while samples of the different classes in different modalities represent negative samples. The softmax function normalises the representations of positive and negative samples, and the similarity ranges from 0 to 1. The intra-modal and inter-modal contrastive loss is defined as follows:
\begin{equation}
	\mathcal{L}_{I M C L}=-\mathbb{E}_s\left[\frac{\sum_{i=1}^{2N}\left(\mu^{\mathcal{M}}\right)^T \chi_i^{\mathcal{M}}}{\sum_{i=1}^{2 N}\left(\mu^{\mathcal{M}}\right)^T \chi_i+\sum_{j=1}^{2 M}\left(\mu^{\mathcal{M}}\right)^T \delta_j^\mathcal{M}}\right]
\end{equation}
where $\mu^\mathcal{M}$ denotes the anchor embedded representation, $N$ denotes the number of positive samples, $M$ denotes the number of negative samples, $\chi_i^\mathcal{M}$ and $\delta_j^\mathcal{M}$ denote the embedded representations of positive and negative samples, respectively. It is noted that $\chi_i^\mathcal{M}$ and $\delta_j^\mathcal{M}$ are the same modality and different classes. 

For ICCL, regularisation terms are introduced to strengthen the similarity between positive sample pairs, which reduces the similarity between negative samples. The formula is defined as follows:

\begin{equation}
	\begin{aligned}
		\mathcal{L}_{I M C L}^R= & \mathbb{E}_S\left[\frac{1}{2 N} \sum_{i=1}^N\left\|\left(\mu^{\mathcal{M}}\right)^T \delta_j-\beta\right\|^2\right] \\
		& \tilde{\mathcal{L}}_{I M C L} = \mathcal{L}_{I M C L}+\mathcal{L}_{I M C L}^R
	\end{aligned}
\end{equation}
where $\mathcal{L}_{I M C L}$ is the regularization loss for IMCL. IMCL encourages high similarity between samples of the same class in the same modality, which forces low similarity between samples of the same class in different modalities. With IMCL, the model can learn intra-modal and inter-modal complementary semantic information.

\subsection{ Emotion Inference Subnetwork}
After the multimodal feature vectors pass through the matching attention layer, each contextual utterance can be represented as a multimodal fusion vector $z^f$. We use a multi-layer perceptron deep neural network called the Emotion Inference Subnetwork $\mathcal{G}_s$ with weights $W$ conditioned on $z^f$. The multi-layer perceptron (MLP) consists of two fully connected layers with ReLU activation functions and connects them to a decision layer. The maximum likelihood function of the Emotion Inference Subnetwork $\mathcal{G}_s$ is defined as follows, where $\varphi$ is the label of emotion prediction:

\begin{equation}
	\mathcal{L}_{CLS}=\arg \max _{\varphi} p\left(\varphi \mid z^f ; W\right)=\arg \max _{\varphi} \mathcal{G}_s\left(z^f, W\right)
\end{equation}
where $\mathcal{L}_{CLS}$ is emotion classification loss for the model. The smaller $\mathcal{L}_{CLS}$, the better the emotion classification effect.

\begin{algorithm}[]
		\caption{Adversarial Representation with Intra-Modal and Inter-Modal Graph Contrastive Learning (AR-IIGCN)} 
	\label{alg:alg}
	\textbf{Input:} The length of the training dataset $L$, the multi-modal feature vectors $\xi_u, \xi_v, \xi_a$,    
	the batch size for model training $\mathcal{B}$,            
	iterations for model training $K$,              
	degree matrix $A$ of the relationship graph,            
	and the modal margin  $\beta$.                                                                                                                                                                                                                                                                                                                                                                 \\
	\textbf{Output:} The predicted labels $\hat{y}$. 
    	\begin{algorithmic}[1]

    	\STATE Initalize the model weight of AR-IIGCN.
        \STATE Initalize a collection of predicted emotion labels  $\hat{y}$.   
  
       \FOR{ epoch = $1,2,\ldots, K$ }
       \FOR{ $i=1,2,\ldots, L/\mathcal{B}$}
	\STATE Sample    $\xi=\left\{\xi_u^i, \xi_v^i, \xi_a^i\right\}_{i=1}^{\mathcal{B}}$.
	\STATE $\hat{\xi}_u^i, \hat{\xi}_v^i, \hat{\xi}_a^i=F_t\left(\xi_u^i\right), F_v\left(\xi_v^i\right), F_a\left(\xi_a^i\right)$.
	\STATE $\tilde{\xi}_u^i=D_T\left(G_T\left(\hat{\xi}_u^i\right)\right)$. 
	\STATE $\tilde{\xi}_v^i, \tilde{\xi}_a^i = D_v\left(G_v\left(\hat{\xi}_v^i\right)\right), D_A\left(G_A\left(\hat{\xi}_a^i\right)\right)$.
	\STATE Update generator and discriminator parameters by Eq. (2-7).
	\ENDFOR
	\ENDFOR  

    		\FOR {epoch $1,2,\ldots, K$}
    		\FOR{$i=1,2,\ldots, L/\mathcal{B}$}
    		\STATE Sample $\hat{\xi}=\left\{\hat{\xi}_u^i, \hat{\xi}_v^i, \hat{\xi}_a^i\right\}_{i=1}^{\mathcal{B}}$ .
    		\STATE $\omega_{i j}^M=\operatorname{softmax}\left(W_{\vartheta_1}^{\mathcal{M}}\left(\operatorname{GELU}\left(W_{\vartheta_2}^{\mathcal{M}}\left[\tilde{\xi}_i^{\mathcal{M}} \oplus \tilde{\xi}_j^{\mathcal{M}}\right]\right)\right)\right)$.
    		\STATE             Obtain the node representation $\psi^t, \psi^a, \psi^v$ by Eq. (10). 
    		\STATE             Sample positive and negative samples $\chi^{\mathcal{M}}=\left\{\psi_i^t, \psi_i^a, \psi_i^v\right\}$. 
    		\STATE             Calculate ICCL and IMCL losses by Eq. (12) and Eq. (14). 
    		\STATE             Obtain predicted emotion labels $\hat{y}_i$ via MLP. 
    		\STATE            Update network parameters by Eq. (17). 
    		\ENDFOR 
    		\ENDFOR 
    		\STATE \textbf{Return} the predicted emotion labels $\hat{y}$.                                                                                                                                                                                                                                                                                                                                                                                                                                                                                                                                                                                                                                                            
    	\end{algorithmic}
\end{algorithm}

\subsection{Model Training}
The intra-modal, inter-modal, and intra-class and inter-class contrastive losses are obtained by weighted summation of IMCL and ICCL. The formula is defined as follows:
\begin{equation}
	\mathcal{L}_{\text {hybrid }}=\lambda \tilde{\mathcal{L}}_{I M C L}+(1-\lambda) \tilde{\mathcal{L}}_{I C C L}
\end{equation}

The overall loss for model training is obtained by summing the classification loss and the contrastive loss. The formula for the model training loss is defined as follows:
\begin{equation}
	\mathcal{L}_{\text {overall }}=\mathcal{L}_{C L S}+\mathcal{L}_{\text {hybrid }}
\end{equation}
where $\mathcal{L}_{overall}$ is the overall loss of the model. The smaller $\mathcal{L}_{overall}$ is, the better the training effect of the model is.

The entire inference process of the AR-IIGCN pseudocode is contained in Algorithm 1.

\subsection{Implementation Details}
In this section, we have described the implementation details of the model during training. We have divided the benchmark dataset into three parts. The first part is the training set for model training, and the second part is the validation set for updating the network parameters. The third part is the test set for evaluating the emotional prediction effect of the model. In addition, the ratio of the training set, test set, and validation set is 8:1:1. The experimental environment is the Windows 10 operating system, and the hardware driver is a computer with Nvidia RTX 3090. It is used Python 3.8 and Pytorch 1.9.1 versions to complete the construction of deep learning algorithms. To evaluate the effective convergence of the model, this paper is used the highly stable Adam algorithm [10] to optimize the network parameters. In addition, it is set as the epoch size to 60, batch size to 32, learning rate to 0.0005, dropout to 0.5, and weight decay coefficient to 0.00001.

\section{Experiments}

\subsection{Benchmark Dataset Used}
The MELD \cite{oria-etal-2019-meld} and IEMOCAP \cite{busso2008iemocap} multimodal conversation datasets are often used for comparative experiments in MERC. Table 1 shows the distribution of MELD and IEMOCAP datasets in each emotion category. We introduce the situation of the two datasets as follows.

The Interactive Emotional Dyadic Motion Capture Database (IEMOCAP) contains three modalities, namely video, audio, and text. Therefore, IEMOCAP is a multimodal dataset, and the use of multimodal emotion recognition methods can enhance the prediction effect of the model. A total of 10 actors and actresses are included in the IEMOCAP dataset, and they communicate in an interactive way. For each conversation, it is annotated by multiple sentiment experts, avoiding the subjectivity of human annotation. In addition, the IEMOCAP dataset contains a total of six emotions, namely "sad", "happy", "angry", "neutral", "frustrated" and "excited".

The Multi-modal EmotionLines Dataset (MELD) is also a multi-modal dataset whose corpus consists of dialogues from the TV series Friends. Similar to the IEMOCAP dataset, each conversation is also annotated by multiple sentiment experts. In addition, the MELD dataset contains a total of seven emotions, namely "disgust", "anger", "joy", "fear", "sadness", "neutral", and "surprise".

\subsection{Evaluation Metrics}
To compare the emotion recognition effect of our algorithm and other baseline algorithms, we use four evaluation indicators: 1) Accuracy; 2) F1; 3) Weighted average accuracy (WAA); 4) Weighted average F1 (WAF1). We define these four indicators as follows:

1) $Accuracy_j$ is the prediction accuracy of the model on the $j$-th emotion category, and its formula is defined as follows:
\begin{equation}
	\operatorname {Accuracy}_j=\frac{\sum_{k=1}^M P_j^k}{\sum_{i=1}^N T_j^i}
\end{equation}
where $M$ is the number of samples that the model correctly predicts on the $j$-th sentiment. $N$ is the total number of samples included in the $j$-th sentiment. $T_j^i$ represents the $i$-th sample on the $j$-th sentiment. $P_j^k$ means that the model predicts the $k$-th sample correctly on the jth category of emotion, and $P_j^k \in [0,1]$. Relatively speaking, the higher the $Accuracy_j$ value, the higher the confidence of the model prediction.

2) The value of $F1_j$ represents the f1-score predicted by the model on the $j$-th emotion category, and its formula is defined as follows:
\begin{equation}
	F 1_j=\frac{2 \times \operatorname{Recall}\left(P_{T P}^j, P_{F P}^j\right) \times \operatorname{Precision}\left(P_{T P}^j, P_{F N}^j\right)}{\operatorname{Recall}\left(P_{T P}^j, P_{F P}^j\right)+\operatorname{Precision}\left(P_{T P}^j, P_{F N}^j\right)}
\end{equation}

and
\begin{equation}
	\begin{aligned}
		\operatorname{Precision}\left(P_{T P}^j, P_{F N}^j\right) &=\frac{\left|P_{T P}^j\right|}{\left|E_{T P}^j \cup E_{F N}^j\right|} \\
		\operatorname{Recall}\left(P_{T P}^j, P_{F P}^j\right) &=\frac{\left|P_{T P}^j\right|}{\left|P_{T P}^j \cup P_{F P}^j\right|}
	\end{aligned}
\end{equation}
where, $P_{TP}^j$ represents the number of positive samples predicted on the $j$-th emotion, and $P_{FN}^j$ represents the number of negative samples predicted on the $j$-th emotion. $P_{FP}^j$ represents the number of other emotion categories identified by the model as the jth positive samples. $Precision(P_{TP}^j, P_{FN}^j )$ represents the precision of the model's recognition on the $j$-th type of emotion, and $Recall(P_{TP}^j, P_{FP}^j )$ represents the recall of the model's recognition on the $j$-th type of emotion. The F1 value combines the two metrics of precision and recall. In particular, the higher the F1 value, the higher the confidence of the model's emotion prediction.

3) WAA takes into account the problem of imbalanced sentiment categories and is the weighted average of the prediction accuracy of the model across all categories. The weight of the sample is inversely proportional to the number of samples, and the formula is defined as follows:
\begin{equation}
	W A A=\frac{\sum_{j=1}^\tau T_j * \text {Accuracy}_j}{\sum_{j=1}^\tau \sum_{i=1}^N T_j^i}
\end{equation}
where $\tau$ is the total number of sentiment categories in the dataset. The higher the WAA, the better the comprehensive prediction effect of the model.

4) Like WAA, WF1 also considers the problem of imbalanced sentiment category divisions, which is the weighted average of the model's f1-score across all categories. The weight of the sample is inversely proportional to the number of samples, and the formula is defined as follows:
\begin{equation}
	W F 1=\frac{\sum_{j=1}^\tau T_j * F 1_j}{\sum_{j=1}^\tau \sum_{i=1}^N T_j^i}
\end{equation}
Specifically, the higher the WF1, the higher the prediction confidence of the model.

\subsection{Baseline Models}
We do extensive comparative experiments on two popular datasets to count the emotion recognition effect of the model proposed in this paper. Some recent comparison algorithms are described below:

\textbf{TextCNN:} The TextCNN proposed by Kim \cite{kim-2014-convolutional} uses Convolutional Neural Networks (CNN) for emotion recognition of dialogues. TextCNN exploits the local attention mechanism of convolution kernels to extract contextual utterances with emotional polarity in texts. However, TextCNN cannot model the context of long-range dependencies and can only model unimodal features.

\textbf{bc-LSTM:} The bidirectional LSTM (bc-LSTM) proposed by Poria et al. \cite{poria-etal-2017-context} can not only model long-range contextual dependencies, but also extract contextual information in two opposite directions, thereby eliminating word ambiguity. However, bc-LSTM does not model speaker relations.

\textbf{DialogueRNN:} The DialogueRNN proposed by Ghosal et al. consists of three gated recurrent units (i.e., global GRU, party GRU and emotion GRU), which are able to distinguish between speakers.

\textbf{DialogueGCN:} DialogueGCN proposed by Ghosal et al. \cite{ghosal-etal-2019-dialoguegcn} is the first to use graph convolutional neural networks (GCNs) to model speaker relations. DialogueGCN simulates the dialogue relationship between speakers by constructing a fully connected directed graph, which can fuse the contextual semantic information and the semantic information of the dialogue relationship between speakers. However, the fully connected graph constructed by DialogueGCN may introduce noisy information.

\textbf{CTNet:} The Conversational Transformer Network proposed by Lian et al. \cite{9316758} comprehensively considers intra-modal and inter-modal modeling, and captures long-range contextual information by using a cross-modal Conversational Transformer architecture.

\textbf{LR-GCN:} The LR-GCN proposed by Ren et al. \cite{9556142} not only utilizes GCN to model the relationship between speakers, but also utilizes a multi-head attention mechanism to model the latent relationship between utterances. In addition, to speed up the convergence of the model, LR-GCN also introduces a residual structure to transfer more gradient information. LR-GCN has achieved good experimental results.

\begin{table*}[!t]
	\renewcommand\arraystretch{1.5}
	\setlength{\tabcolsep}{12pt}
	\caption{Experimental results with our method and other baseline on IEMOCAP dataset. The best result in each column is in bold. Average(w) represents the weighted average.}
	\begin{tabular}{l|ccccccc}
		\hline
		\multirow{3}{*}{Methods} & \multicolumn{7}{c}{IEMOCAP}                                                              \\ \cline{2-8} 
		& Happy      & Sad        & Neutral    & Angry      & Excited    & Frustrated & Average(w) \\ \cline{2-8} 
		& Acc.  F1   & Acc.  F1   & Acc.  F1   & Acc.  F1   & Acc.  F1   & Acc.  F1   & WAA  WF1   \\ \hline
		TextCNN                      & 27.73  29.81 & 57.14  53.83 & 34.36  40.13 & 61.12  52.47 & 46.11  50.09 & 62.94  55.78 & 48.93  48.17 \\
		bc-LSTM                  & 29.16  34.49 & 57.14  60.81 & 54.19  51.80 & 57.03  56.75 & 51.17  57.98 & 67.12  58.97 & 55.23  54.98 \\
		bc-LSTM+Att & 30.56 35.63 & 56.73 62.09 & 57.55 53.00 & {59.41} 59.24 & 52.84 58.85 & 65.88 {59.41} & 56.32 56.19  \\
		DialogueRNN              & 25.63  33.11 & 75.14  78.85 & 58.56  59.24 & 64.76  65.23 & 80.27  71.85 & 61.16  58.97 & 63.42  62.74 \\
		DialogueGCN              & 40.63  42.71 & \textbf{89.14  84.45} & 61.97  63.54 & 67.51  64.14 & 65.46  63.08 & 64.13  66.90 & 65.21  64.14 \\
		CT-Net                & 47.97  51.36 & 78.01  79.94 & \textbf{69.08  65.82} & 72.98  67.21 & \textbf{85.35}  78.74 & 52.27  58.83 & 68.01  67.55   \\
		LR-GCN                   & 54.24  55.51 & 81.67  79.14 & 59.13  63.84 & 69.47  \textbf{69.02} & 76.37  74.05 & \textbf{68.26  68.91} & 68.52  68.35 \\ 
		AR-IIGCN & \textbf{71.88 69.96} & 74.64 81.58 &67.25 63.80 & \textbf{73.79} 68.37 & 82.66 \textbf{79.15} &{60.45 63.95} & \textbf{70.46 70.36} \\ \hline
	\end{tabular}
\end{table*}

\begin{table*}[!t]
	\renewcommand\arraystretch{1.5}
	\caption{Experimental results with our method and other baseline on MELD dataset. The best result in each column is in bold. Average(w) represents the weighted average.}
	\setlength{\tabcolsep}{8.6pt}{
		\begin{tabular}{l|cccccccc}
			\hline
			\multirow{3}{*}{Methods} & \multicolumn{8}{c}{MELD}                                                                            \\ \cline{2-9} 
			& Neutral     & Surprise    & Fear     & Sadness    & Joy        & Disgust  & Anger      & Average(w) \\ \cline{2-9} 
			& Acc.  F1    & Acc.  F1    & Acc.  F1 & Acc.  F1   & Acc.  F1   & Acc.  F1 & Acc.  F1   & WAA WF1   \\ \hline
			TextCNN                      & 76.23  74.91  & 43.35  45.51  & 4.63  3.71 & 18.25  21.17 & 46.14  49.47 & 8.91  8.36 & 35.33  34.51 & 56.35  55.01 \\
			bc-LSTM                  & 78.45   73.84 & 46.82   47.71 & 3.84  5.46 & 22.47  25.19 & 51.61  {51.34} & 4.31  5.23 & 36.71  38.44 & 57.51  55.94 \\
			bc-LSTM+Att & 70.45 75.55 & 46.43 46.35 & 0.00 0.00 &21.77 16.27 & 49.30 50.72 & 0.00 0.00 & 41.77 40.71 & 58.51 55.84 \\
			DialogueRNN              & 72.12   73.54 & {54.42}  49.47  & 1.61  1.23 & 23.97  23.83 & 52.01  50.74 & 1.52  1.73 & 41.01  41.54 & 56.12  55.97 \\
			CT-Net                & 75.61   77.45       & 51.32  52.76        & 5.14  \textbf{10.09}     & 30.91  32.56       & 54.31  56.08       & \textbf{11.62  11.27}     & \textbf{42.51}  44.65       & 61.93    60.57 \\
			AR-IIGCN &\textbf{81.10 81.19} & \textbf{56.16 57.24} & \textbf{6.90}  5.06& \textbf{47.41 37.32} & \textbf{65.92 65.92} &{2.94 2.94} &42.28 \textbf{45.22} &\textbf{64.14 64.01}\\ \hline
	\end{tabular}}
\end{table*}

\section{Results and Discussion}

\subsection{Comparison with Baselines}
This paper compares our proposed emotion recognition algorithm AR-IIGCN with other deep learning algorithms. Table 1 and Table 2 show the recognition accuracy and F1 value of all algorithms on each emotion category on the IEMOCAP and MELD datasets, and the average accuracy and F1 value of the model. Experimental results demonstrate the superiority of our algorithm.

\textbf{IEMOCAP:} As shown in Table 1, AR-IIGCN has the best emotion recognition effect on the IEMOCAP dataset, and the WAA and WF1 values are 70.46\% and 70.36\%, respectively. In addition, AR-IIGCN has the highest accuracy rate on the "happy" and "angry" classes, and the highest F1 value on the "happy" and "excited" classes, while the accuracy and F1 value of other categories are slightly lower than other comparison algorithms. The reason is that AR-IIGCN comprehensively considers the heterogeneity of modalities, the intra-modal and inter-modal complementary semantic information, and the intra-class and inter-class differences. The emotion recognition effect of LR-GCN is second, and the values of WAA and WF1 are 68.52\% and 68.35\%, respectively. The reason why LR-GCN is less effective than AR-IIGCN is that it ignores the heterogeneity of modalities, which leads to poor learning effect of subsequent class boundaries. The effects of other algorithms are relatively poor, and they do not consider the heterogeneity of modalities and the intra-class and inter-class differences.

MELD: As shown in Table 2, AR-IIGCN has the best emotion recognition effect on the MELD dataset, and the WAA and WF1 values are 64.14\% and 64.01\%, respectively. In addition, AR-IIGCN has the highest accuracy on the "neutral", "surprise", "fear", "joy" and "sadness" categories, the F1 values on the "neutral", "surprise", "joy", "sadness" and "angry" categories are the highest, while the accuracy and F1 values in other categories are slightly lower than other comparison algorithms. In the "fear" and "disgust" categories, the recognition accuracy and F1 value of AR-IIGCN and other models are low, because the MELD dataset has a serious category imbalance problem.

The analysis of the above experimental results illustrates the superior performance of AR-IIGCN, which can effectively learn the class boundary information of emotions.

\subsection{Importance of the Modalities}
Since different modal features contain different semantic information, we explored the emotion recognition effect of different modal features on the IEMOCAP and MELD datasets. As shown in Table 3, text features perform best in emotion recognition in single-modal experiments, with WA values of 65.4\% and 60.8\%, and WF1 values of 60.8\% and 60.1\% in IEMOCAP and MELD datasets, respectively. We think this is because text is the most direct way for speakers to express their emotions, and it contains the least noisy information. Audio features perform second best for emotion recognition, while video features perform the worst. We think this is because video features contain too much noise information, and it is difficult for the model to extract key information. The emotion recognition effect of the combination of text, audio and video features is the best in all experiments, because the model effectively utilizes the complementary semantic information between modalities. The above experimental phenomena also prove the rationality of our mode fusion layer design.
\begin{table}[htbp]
	\renewcommand\arraystretch{1.5}
	\setlength{\tabcolsep}{5.6mm}{
		\caption{The effect of AR-IIGCN on IEMOCAP and MELD datasets using unimodal features and multimodal features, respectively. T, V, and A represent text, video, and audio modality features. The best result in each column is in bold.}
		\begin{tabular}{c|cccc}
			\hline
			\multirow{2}{*}{Modality} & \multicolumn{2}{c}{IEMOCAP} & \multicolumn{2}{c}{MELD} \\ \cline{2-5} 
			& WA       & WF1              & WA         & WF1         \\ \hline
			T                         & 65.4        & 64.9       & 60.8          & 60.1           \\
			A                         & 62.3        & 62.0                & 58.6          & 57.7           \\
			V & 56.2 & 54.5  & 56.8 & 54.3 \\
			T+A+V                     & \textbf{70.5}        & \textbf{70.4}                & \textbf{64.1}          & \textbf{64.0}           \\ \hline
	\end{tabular}}
\end{table}

\subsection{Effectiveness of Cross-modal Feature Fusion}
In this section, to compare the difference between our proposed multimodal feature fusion method and other methods in multi-modal emotion recognition, we compare our method combining trimodal generative adversarial networks and graph contrastive learning with the other four feature fusion methods.

\textbf{Add:} The Add method combines the feature vectors by summing the multimodal features, which ignores the information interaction between the multimodal features.

\textbf{Concatenate:} The Concatenate method is a splicing operation of multi-modal features, which does not model multi-modal features within and between modalities.

\textbf{TFN:} TFN method models the fusion between multimodal features through tensor outer product operations.

\textbf{LFM:} LFM fuses multimodal features through low-rank tensors.

\begin{table}[htbp]
	\renewcommand\arraystretch{1.5}
	\setlength{\tabcolsep}{3.4mm}{
		\caption{Emotion recognition effects of different multimodal feature fusion methods on IEMOCAP and MELD datasets. We use multi-modal features for each method. The best result in each column is in bold.}
		\begin{tabular}{c|cccc}
			\hline
			\multirow{2}{*}{Methods} & \multicolumn{2}{c}{IEMOCAP} & \multicolumn{2}{c}{MELD} \\ \cline{2-5} 
			& WA       & WF1              & WA         & WF1         \\ \hline
			Add                      & 55.2        & 55.0       & 57.5          & 55.9           \\
			Concatenate              & 58.3        & 57.4               & 58.6         & 57.1           \\
			Tensor Fusion            & 63.2        & 63.0                & 59.6          & 58.7           \\
			Low-rank Fusion            & 63.8        & 63.6                & 60.8          & 59.5           \\
			Cross-modal Fusion(Ours)  &\textbf{70.5}        & \textbf{70.4}                & \textbf{64.1}          & \textbf{64.0}           \\ \hline
	\end{tabular}}
\end{table}

\begin{figure*}
	\centering
	\includegraphics[width=1.01\linewidth]{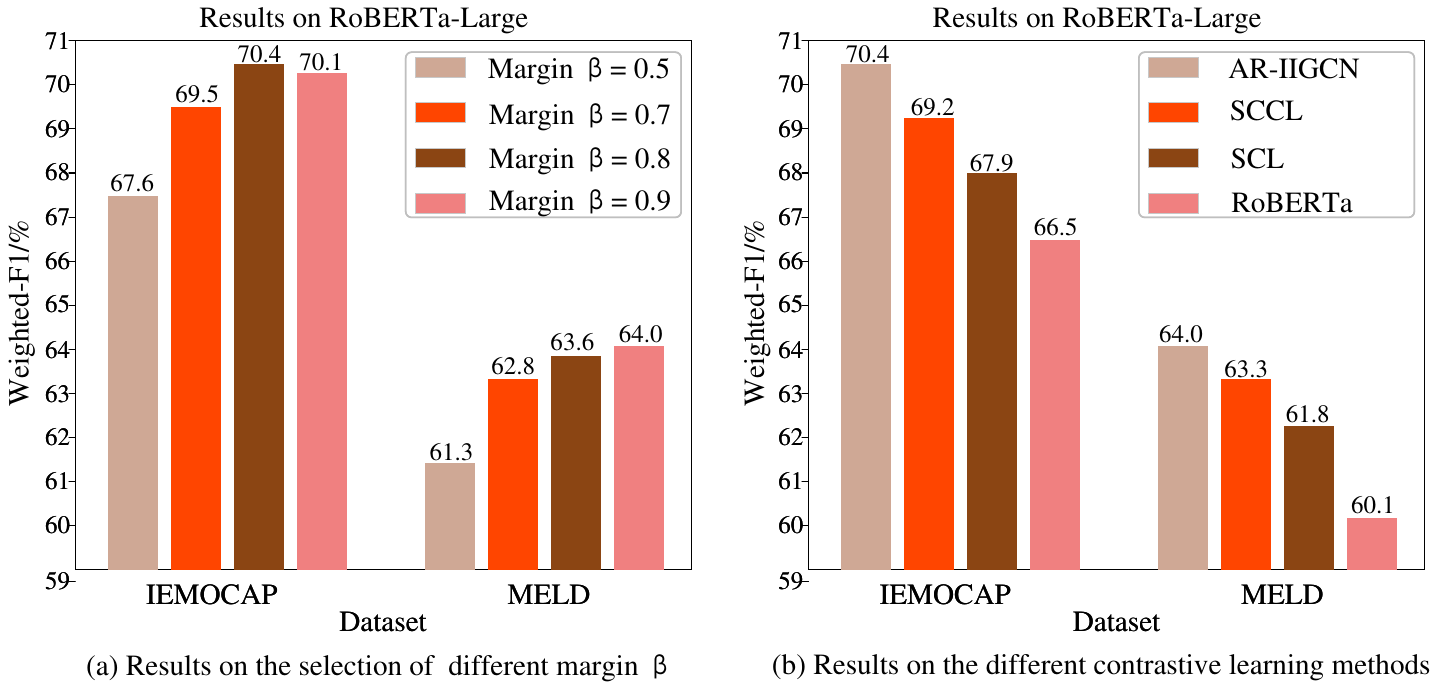}
	\caption{Experimental results on RoBERTa-Large. (a) Effect of different modal margins $\beta$ on model training results. (b) Effect of different contrastive learning methods on model training results.}
	\label{fig:}
\end{figure*}

As shown in Table 4, compared with other fusion methods, our cross-modal fusion method achieves the best emotion recognition performance, with WA values of 70.5\% and 64.1\% and WF1 values of 70.4\% and 64.0\% on IEMOCA and MELD datasets, respectively. Specifically, our method improves the WA value by 15.3\% and 6.6\% and the WF1 value by 15.4\% and 8.1\% over the Add method on the IEMOCAP and MELD datasets, respectively. This is because the Add method cannot eliminate the heterogeneity among modalities and cannot utilize complementary semantic information between modalities. Compared with the Concatenate method, our method improves the WA value by 12.2\% and 6.9\%, and the WF1 value by 5.5\% and 6.9\%, respectively. Similar to the Add method, the Concatenate method cannot take advantage of complementary semantic information between modals. Compared with the Add and Concatenate methods, the Tensor Fusion and Low-rank Fusion methods have significantly improved results, because they utilize complementary semantic information between modalities, and the Low-rank Fusion method can reduce redundant information between modalities . However, they cannot eliminate the heterogeneity between modalities. The above experiments illustrate the superiority of our designed cross-modal approach and the necessity of eliminating modality heterogeneity.

\begin{table}[htbp]
	\renewcommand\arraystretch{1.25}
	\setlength{\tabcolsep}{2.8mm}{
		\caption{The results of the equal parameters experiment on IEMOCAP and MELD datasets. The parameters of methods with $\diamond$ are incremented to be the same as methods with AR-IIGCN. The best result in each column is in bold.}
		\label{table3}
		\begin{tabular}{l|ccccc}
			\hline
			\multirow{2}{*}{Method} & \multicolumn{1}{c}{\multirow{2}{*}{Params}} & \multicolumn{2}{c}{IEMOCAP}                           & \multicolumn{2}{c}{MELD}                              \\ \cline{3-6} 
			& \multicolumn{1}{c}{}                       & \multicolumn{1}{c}{WAA}  & \multicolumn{1}{c}{WF1}    & \multicolumn{1}{c}{WAA}  & \multicolumn{1}{c}{WF1}    \\ \hline
			bc-LSTM                 & 0.53M                                      & 55.2                     & 54.9                     & 57.1                     & 56.4                     \\
			bc-LSTM$^\diamond$                 & 14.68M                                      & 52.9                     & 52.7                     & 53.3                     & 52.9                                          \\ \hline
			DialogueRNN & 13.19M & 63.4 & 62.7 & 56.1 & 56.0 \\
			DialogueRNN$^\diamond$ & 14.68M & 62.7 & 62.2 & 54.8 & 54.1 \\ \hline
			AR-IIGCN                 & 14.68M       &    \textbf{70.5}      & \textbf{70.4}  & \textbf{64.1}        & \textbf{64.0}           \\ \hline
	\end{tabular}}
\end{table}

\subsection{Equal Parameter Experiments}
To illustrate that our method AR-IIGCN does not improve the performance of the model due to the increase in the number of parameters, but the performance improvement caused by the architecture design of the model, we conducted experiments with equal parameters on the IEMOCAP and MELD datasets. As shown in Table 5, the WA and WF1 values of emotion recognition of the bc-LSTM and DialogueRNN models decreased when the number of parameters increased. In addition, in the process of observing the model training, we find that the model is more prone to overfitting as the number of parameters increases. Therefore, the above phenomenon shows that our model architecture outperforms existing emotion recognition algorithms.

\subsection{Comparison of Modality Margin $\beta$}
Since the modal margin $\beta$ is a hyperparameter in this paper, we have conducted extensive experiments to verify the effect of different margins $\beta$ on emotion recognition. As shown in Fig. 4(a), on the IEMOCAP dataset, our model performs best in emotion recognition with a margin $\beta = 0.8$, with a WF1 value of 70.4\%. When the margin is too small (e.g., $\beta=0.5/0.7$), the contrastive learning ability of the model is poor resulting in still large modal gaps. On the contrary, if the margin is too large (e.g., $\beta=0.9$), complementary semantic information between different modalities may be lost. On the MELD dataset, our model performs best in emotion recognition with a margin $\beta = 0.9$, with a WF1 value of 70.4\%. Similar to on the IEMPCAP dataset, too small margins can lead to large modal gaps. Therefore, choosing a good margin has an important impact on the training effect of the model. 

\begin{figure}
	\centering
	\includegraphics[width=1.1\linewidth]{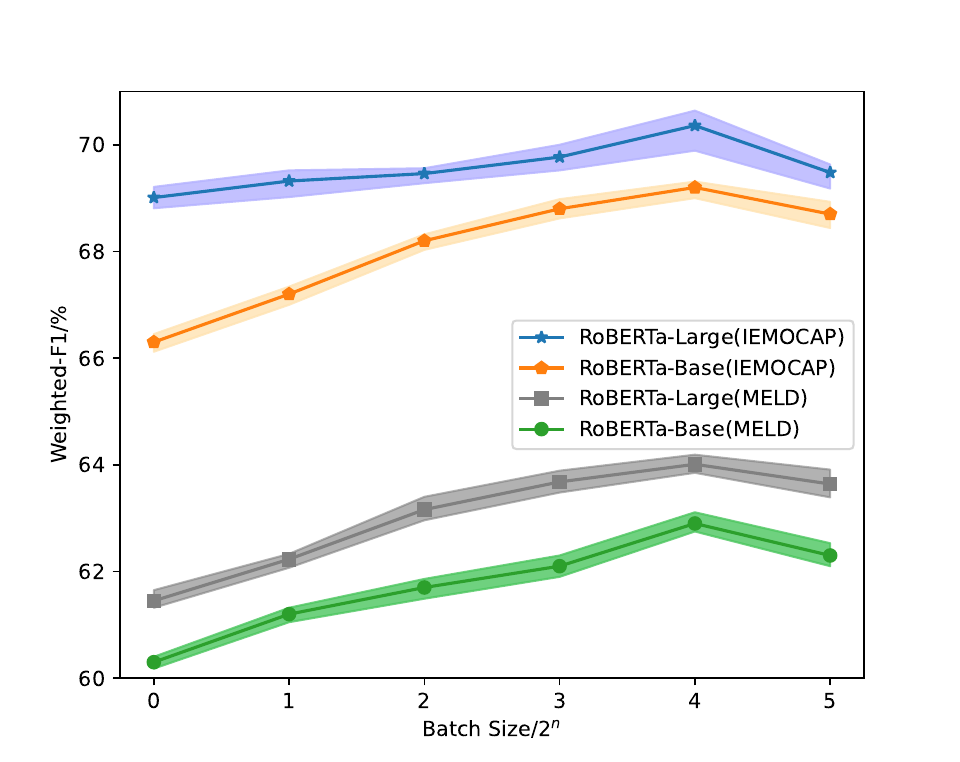}
	\caption{We use different batch sizes with RoBERTa-Large to verify the stability experiments on IEMOCAP and MELD datasets.}
	\label{fig:}
\end{figure}

\subsection{Comparison of Contrastive Learning Methods}
To explore the effectiveness of our designed graph contrastive learning mechanism, we compared different contrastive learning methods, i.e., Supervised Contrastive Learning (SCL), Supervised Cluster-level Contrastive Learning (SCCL).

As shown in Fig. 4(b), we use RoBERTa-large as our text encoder to obtain rich context semantic information. For the results on different comparative learning methods, SCL achieves a 1.4\% improvement over the RoBERTa baseline on the IEMOCAP dataset and a 1.7\% improvement on the MELD dataset. The performance improvement of SCCL on the IEMOCAP and MELD datasets is higher than SCL, achieving 2.7\% and 3.2\% improvements respectively compared to the RoBERTa baseline. AR-IIGCN has the highest performance improvement on the IEMOCAP and MELD datasets, achieving 3.9\% and 3.9\% improvements compared to the RoBERTa baseline, respectively.

The above experimental phenomena illustrate the effectiveness of our designed intra-modal and inter-modal, and intra-class and inter-class contrastive learning mechanism.

\subsection{Batch Size Stability}
We use different batch sizes to verify the stability of model training on IEMOCAP and MELD datasets. As shown in Fig. 5, We set the batch size to range from $2^0=1$ to $2^5=32$. According to the experimental results, the model has the best emotion classification effect when the batch size is 16. When each training step sets a small batch size (i.e., when only a small number of samples are used), the model cannot extract effective features in different modalities, and their contrastive learning effect will be relatively poor.

\begin{figure*}
	\centering
	\includegraphics[width=1.01\linewidth]{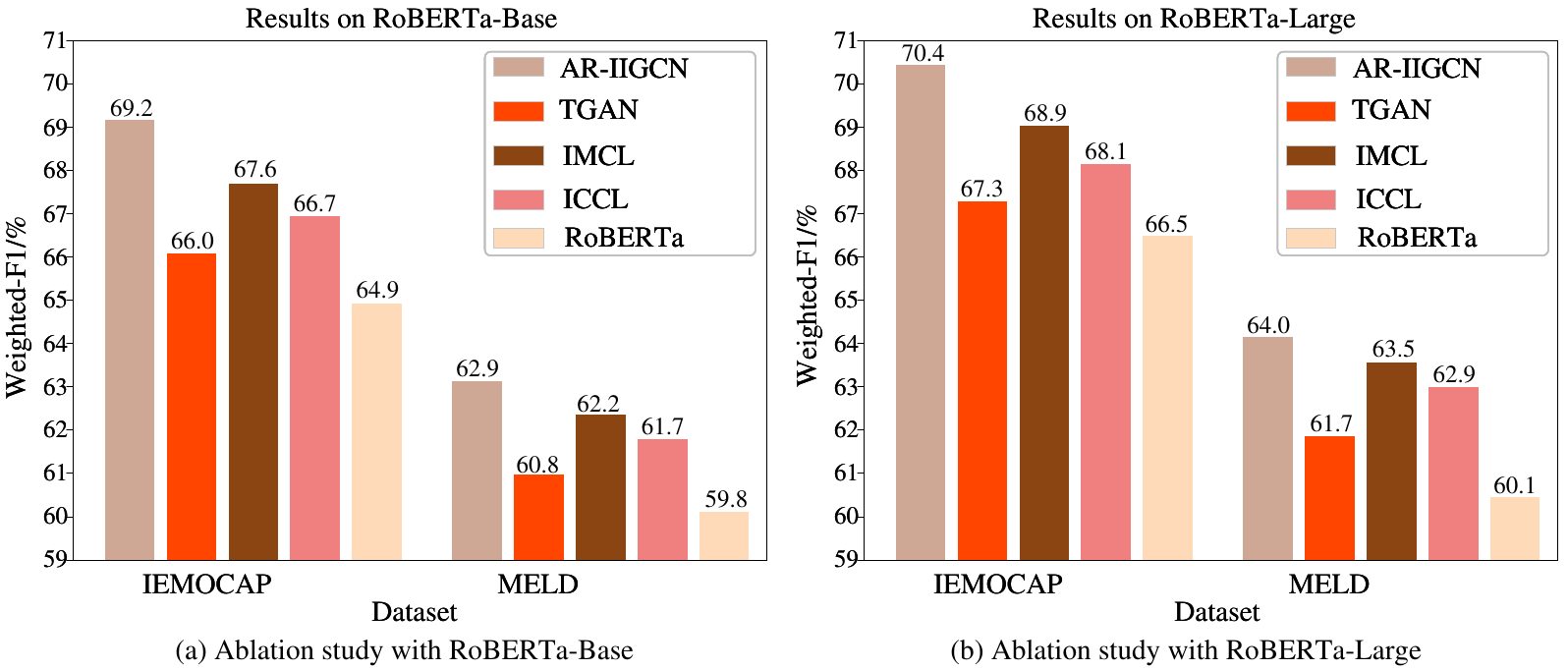}
	\caption{Ablation experiments on IEMOCAP and MELD datasets. (a) We use RoBERTa-Base as a text encoder to explore the impact of TGAN, IMCL, and ICCL on model training. (b) We use RoBERTa-Large as a text encoder to explore the impact of TGAN, IMCL, and ICCL on model training.}
	\label{fig:ar-iigcn}
\end{figure*}

\subsection{Extended Research}
To verify the scalability of our fusion and contrast mechanism to other multimodal studies, we apply our method to the task of multimodal humor detection. As shown in Table 6, C-MFN (C) means using only contextual information without punchlines. C-MFN (P) means using only punchlines with no contextual information, and C-MFN represents using punchlines and contextual information. We embed our TGAN, IMCL, and ICCL mechanisms into the C-MFN method, and experimental results show that our method outperforms the C-MFN method in any combination of modalities. Specifically, we use accuracy as the evaluation metric for humor detection, and our method can achieve improvements ranging from 1.48\% to 7.57\%. Experimental results show that our method can be applied not only to multimodal emotion recognition tasks, but also to other multimodal tasks.

\begin{table}[htbp]
	\renewcommand\arraystretch{1.25}
	\setlength{\tabcolsep}{3.3mm}{
		\caption{Experimental results of C-MFN method on the UR-FUNNY dataset for the humor detection task. C-MFN (C) means using only contextual information without punchlines. C-MFN (P) means using only punchlines with no contextual information. $\ast$ means the method equipped with the TGAN, IMCL, and ICCL module. The best result is highlighted in bold.}
		\label{table7}
		\begin{tabular}{lccccc}
			\hline
			\multicolumn{6}{c}{UR-FUNNY}                                                 \\ \hline
			\multicolumn{1}{l|}{Modality}        & T     & A+V   & T+A   & T+V   & T+A+V \\ \hline
			\multicolumn{1}{l|}{C-MFN(P)}        & 62.85 & 53.30 & 63.28 & 63.22 & 64.47 \\
			\multicolumn{1}{l|}{C-MFN(C)}        & 57.96 & 50.23 & 57.78 & 57.99 & 58.45 \\
			\multicolumn{1}{l|}{C-MFN}           & 64.44 & 57.99 & 64.47 & 64.22 & 65.23 \\
			\multicolumn{1}{l|}{C-MFN(P)$^\ast$} & \textbf{67.19} & \textbf{60.87} & \textbf{67.89} & \textbf{68.36} & \textbf{68.97} \\
			\multicolumn{1}{l|}{C-MFN(C)$^\ast$} & \textbf{61.86} & \textbf{55.36} & \textbf{59.26} & \textbf{60.11} & \textbf{62.49} \\
			\multicolumn{1}{l|}{C-MFN$^\ast$}    & \textbf{68.69} & \textbf{62.06} & \textbf{67.68} & \textbf{67.92} & \textbf{68.28} \\ \hline
	\end{tabular}}
\end{table}

\subsection{Ablation Study}
To verify the rationality of our module design, we use RoBERTa-Base and RoBERTa-Large as our text encoders to conduct ablation experiments. As shown in Fig. 6(b), for the results on RoBERTa-Large, AR-IIGCN achieves the best experimental results with WF1 values of 70.4\% and 64.0\% on the IEMOCAP and MELD datasets, respectively. The emotional effect of RoBERTa-Large with IMCL is second, and the WF1 values are 68.9\% and 63.5\%, respectively. The emotional effect of RoBERTa-Large with ICCL is worse than RoBERTa-Large with IMCL, and the WF1 values are 68.1\% and 62.9\% respectively. The emotional effect of RoBERTa-Large with TGAN is only slightly better than the RoBERTa-Large baseline, with WF1 values of 67.3\% and 61.7\%, respectively. The experimental results show that the intra-modal and inter-modal contrastive learning is the most critical for the training of the model, which is beneficial for the model to fuse complementary multi-modal semantic information. Intra-class and inter-class contrastive learning is also important for the training of the model, which facilitates the class boundary learning of the model. Removing the heterogeneity of modalities is the basis for subsequent model learning.

As shown in Fig. 6(a), for the results on RoBERTa-Base,  Similar conclusions are drawn from the results of RoBERTa-Base. In addition, the emotion recognition effect of RoBERTa-Large is better than RoBERTa-Base.

\section{Conclusion and Future Work}
In this paper, we propose a novel Adversarial Representation with Intra-Modal and Inter-Modal Graph Contrastive Learning for Multimodal Emotion Recognition (AR-IIGCN) model, which enables cross-modal feature fusion, intra-modal and inter-modal contrasting representation learning, and intra-class and inter-class representation learning. Specifically, we firstly introduce a cross-modal feature fusion method based on adversarial learning to eliminate the heterogeneity among different modalities. Secondly, to comprehensively consider the relationship between intra-modality and inter-modality and the relationship between intra-class and inter-class, we design a novel graph contrastive learning architecture to enhance the representation ability of nodes by increasing the distance between different emotion labels of the same modality and shrinking the distance between the same emotion of different modalities. Finally, we use a multi-layer perceptron (MLP) for emotion classification.

In future work, we consider using diffusion models for feature fusion across modalities to generate fused features that contain more semantic information. In addition, we will also consider transferring our method to other multimodal tasks.

\section{Ackonwledgments}
This work is supported by National Natural Science Foundation of China	(Grant No. 69189338), Excellent Young Scholars of Hunan Province of China	(Grant No. 20B625), and Changsha Natural Science Foundation (Grant No.	kq2202294).

\bibliographystyle{IEEEtran}
\bibliography{refs}

\begin{thebibliography}{10}
\providecommand{\url}[1]{#1}
\csname url@samestyle\endcsname
\providecommand{\newblock}{\relax}
\providecommand{\bibinfo}[2]{#2}
\providecommand{\BIBentrySTDinterwordspacing}{\spaceskip=0pt\relax}
\providecommand{\BIBentryALTinterwordstretchfactor}{4}
\providecommand{\BIBentryALTinterwordspacing}{\spaceskip=\fontdimen2\font plus
\BIBentryALTinterwordstretchfactor\fontdimen3\font minus
  \fontdimen4\font\relax}
\providecommand{\BIBforeignlanguage}[2]{{%
\expandafter\ifx\csname l@#1\endcsname\relax
\typeout{** WARNING: IEEEtran.bst: No hyphenation pattern has been}%
\typeout{** loaded for the language `#1'. Using the pattern for}%
\typeout{** the default language instead.}%
\else
\language=\csname l@#1\endcsname
\fi
#2}}
\providecommand{\BIBdecl}{\relax}
\BIBdecl

\bibitem{9358000}
F.~Huang, X.~Li, C.~Yuan, S.~Zhang, J.~Zhang, and S.~Qiao,
  ``Attention-emotion-enhanced convolutional lstm for sentiment analysis,''
  \emph{IEEE Transactions on Neural Networks and Learning Systems}, vol.~33,
  no.~9, pp. 4332--4345, 2022.

\bibitem{ai2023gcn}
W.~Ai, Y.~Shou, T.~Meng, and K.~Li, ``Der-gcn: Dialogue and event
  relation-aware graph convolutional neural network for multimodal dialogue
  emotion recognition,'' \emph{arXiv preprint arXiv:2312.10579}, 2023.

\bibitem{meng2023deep}
T.~Meng, Y.~Shou, W.~Ai, N.~Yin, and K.~Li, ``Deep imbalanced learning for
  multimodal emotion recognition in conversations,'' \emph{arXiv preprint
  arXiv:2312.06337}, 2023.

\bibitem{shou2023comprehensive}
Y.~Shou, T.~Meng, W.~Ai, N.~Yin, and K.~Li, ``A comprehensive survey on
  multi-modal conversational emotion recognition with deep learning,''
  \emph{arXiv preprint arXiv:2312.05735}, 2023.

\bibitem{khare2020time}
S.~K. Khare and V.~Bajaj, ``Time--frequency representation and convolutional
  neural network-based emotion recognition,'' \emph{IEEE transactions on neural
  networks and learning systems}, vol.~32, no.~7, pp. 2901--2909, 2020.

\bibitem{zhao2022cauain}
W.~Zhao, Y.~Zhao, and X.~Lu, ``Cauain: Causal aware interaction network for
  emotion recognition in conversations,'' in \emph{Proceedings of the
  Thirty-First International Joint Conference on Artificial Intelligence,
  IJCAI}.\hskip 1em plus 0.5em minus 0.4em\relax Morgan Kaufmann, 2022, pp.
  4524--4530.

\bibitem{zadeh2017tensor}
A.~Zadeh, M.~Chen, S.~Poria, E.~Cambria, and L.-P. Morency, ``Tensor fusion
  network for multimodal sentiment analysis,'' in \emph{Proceedings of the 2017
  Conference on Empirical Methods in Natural Language Processing}, 2017, pp.
  1103--1114.

\bibitem{liu2018efficient}
Z.~Liu, Y.~Shen, V.~B. Lakshminarasimhan, P.~P. Liang, A.~B. Zadeh, and L.-P.
  Morency, ``Efficient low-rank multimodal fusion with modality-specific
  factors,'' in \emph{Proceedings of the 56th Annual Meeting of the Association
  for Computational Linguistics (Volume 1: Long Papers)}, 2018, pp. 2247--2256.

\bibitem{shou2023czl}
Y.~Shou, W.~Ai, T.~Meng, and K.~Li, ``Czl-ciae: Clip-driven zero-shot learning
  for correcting inverse age estimation,'' \emph{arXiv preprint
  arXiv:2312.01758}, 2023.

\bibitem{shou2023graph}
Y.~Shou, W.~Ai, and T.~Meng, ``Graph information bottleneck for remote sensing
  segmentation,'' \emph{arXiv preprint arXiv:2312.02545}, 2023.

\bibitem{meng2023multi}
T.~Meng, Y.~Shou, W.~Ai, J.~Du, H.~Liu, and K.~Li, ``A multi-message passing
  framework based on heterogeneous graphs in conversational emotion
  recognition,'' \emph{Neurocomputing}, p. 127109, 2023.

\bibitem{ying2021prediction}
R.~Ying, Y.~Shou, and C.~Liu, ``Prediction model of dow jones index based on
  lstm-adaboost,'' in \emph{2021 International Conference on Communications,
  Information System and Computer Engineering (CISCE)}.\hskip 1em plus 0.5em
  minus 0.4em\relax IEEE, 2021, pp. 808--812.

\bibitem{shou2022object}
Y.~Shou, T.~Meng, W.~Ai, C.~Xie, H.~Liu, and Y.~Wang, ``Object detection in
  medical images based on hierarchical transformer and mask mechanism,''
  \emph{Computational Intelligence and Neuroscience}, vol. 2022, 2022.

\bibitem{shou2022conversational}
Y.~Shou, T.~Meng, W.~Ai, S.~Yang, and K.~Li, ``Conversational emotion
  recognition studies based on graph convolutional neural networks and a
  dependent syntactic analysis,'' \emph{Neurocomputing}, vol. 501, pp.
  629--639, 2022.

\bibitem{hu2021mmgcn}
J.~Hu, Y.~Liu, J.~Zhao, and Q.~Jin, ``Mmgcn: Multimodal fusion via deep graph
  convolution network for emotion recognition in conversation,'' in
  \emph{Proceedings of the 59th Annual Meeting of the Association for
  Computational Linguistics and the 11th International Joint Conference on
  Natural Language Processing (Volume 1: Long Papers)}, 2021, pp. 5666--5675.

\bibitem{liu2023multi}
S.~Liu, P.~Gao, Y.~Li, W.~Fu, and W.~Ding, ``Multi-modal fusion network with
  complementarity and importance for emotion recognition,'' \emph{Information
  Sciences}, vol. 619, pp. 679--694, 2023.

\bibitem{kosti2017emotion}
R.~Kosti, J.~M. Alvarez, A.~Recasens, and A.~Lapedriza, ``Emotion recognition
  in context,'' in \emph{Proceedings of the IEEE Conference on Computer Vision
  and Pattern Recognition}.\hskip 1em plus 0.5em minus 0.4em\relax IEEE, 2017,
  pp. 1667--1675.

\bibitem{zhang2020emotion}
J.~Zhang, Z.~Yin, P.~Chen, and S.~Nichele, ``Emotion recognition using
  multi-modal data and machine learning techniques: A tutorial and review,''
  \emph{Information Fusion}, vol.~59, pp. 103--126, 2020.

\bibitem{lee2019context}
J.~Lee, S.~Kim, S.~Kim, J.~Park, and K.~Sohn, ``Context-aware emotion
  recognition networks,'' in \emph{Proceedings of the IEEE/CVF International
  Conference on Computer Vision}.\hskip 1em plus 0.5em minus 0.4em\relax IEEE,
  2019, pp. 10\,143--10\,152.

\bibitem{9839579}
Z.~Lian, B.~Liu, and J.~Tao, ``Pirnet: Personality-enhanced iterative
  refinement network for emotion recognition in conversation,'' \emph{IEEE
  Transactions on Neural Networks and Learning Systems}, pp. 1--12, 2022.

\bibitem{poria-etal-2017-context}
S.~Poria, E.~Cambria, D.~Hazarika, N.~Majumder, A.~Zadeh, and L.-P. Morency,
  ``Context-dependent sentiment analysis in user-generated videos,'' in
  \emph{Proceedings of the 55th Annual Meeting of the Association for
  Computational Linguistics (Volume 1: Long Papers)}.\hskip 1em plus 0.5em
  minus 0.4em\relax ACL, 2017, pp. 873--883.

\bibitem{beard-etal-2018-multi}
R.~Beard, R.~Das, R.~W.~M. Ng, P.~G.~K. Gopalakrishnan, L.~Eerens,
  P.~Swietojanski, and O.~Miksik, ``Multi-modal sequence fusion via recursive
  attention for emotion recognition,'' in \emph{Proceedings of the 22nd
  Conference on Computational Natural Language Learning}.\hskip 1em plus 0.5em
  minus 0.4em\relax ACL, 2018, pp. 251--259.

\bibitem{9556142}
M.~Ren, X.~Huang, W.~Li, D.~Song, and W.~Nie, ``Lr-gcn: Latent relation-aware
  graph convolutional network for conversational emotion recognition,''
  \emph{IEEE Transactions on Multimedia}, pp. 1--1, 2021.

\bibitem{nie2020c}
W.~Nie, M.~Ren, J.~Nie, and S.~Zhao, ``C-gcn: correlation based graph
  convolutional network for audio-video emotion recognition,'' \emph{IEEE
  Transactions on Multimedia}, vol.~23, pp. 3793--3804, 2020.

\bibitem{9868807}
S.~Wu, L.~Zhou, Z.~Hu, and J.~Liu, ``Hierarchical context-based emotion
  recognition with scene graphs,'' \emph{IEEE Transactions on Neural Networks
  and Learning Systems}, pp. 1--15, 2022.

\bibitem{9815553}
S.~Qian, D.~Xue, Q.~Fang, and C.~Xu, ``Integrating multi-label contrastive
  learning with dual adversarial graph neural networks for cross-modal
  retrieval,'' \emph{IEEE Transactions on Pattern Analysis and Machine
  Intelligence}, pp. 1--18, 2022.

\bibitem{9609567}
C.-M. Chang and C.-C. Lee, ``Learning enhanced acoustic latent representation
  for small scale affective corpus with adversarial cross corpora
  integration,'' \emph{IEEE Transactions on Affective Computing}, pp. 1--1,
  2021.

\bibitem{li2021contrastive}
M.~Li, B.~Yang, J.~Levy, A.~Stolcke, V.~Rozgic, S.~Matsoukas, C.~Papayiannis,
  D.~Bone, and C.~Wang, ``Contrastive unsupervised learning for speech emotion
  recognition,'' in \emph{ICASSP 2021-2021 IEEE International Conference on
  Acoustics, Speech and Signal Processing (ICASSP)}.\hskip 1em plus 0.5em minus
  0.4em\relax IEEE, 2021, pp. 6329--6333.

\bibitem{kim2021contrastive}
D.~Kim and B.~C. Song, ``Contrastive adversarial learning for person
  independent facial emotion recognition,'' in \emph{Proceedings of the AAAI
  Conference on Artificial Intelligence}, vol.~35, no.~7.\hskip 1em plus 0.5em
  minus 0.4em\relax AAAI, 2021, pp. 5948--5956.

\bibitem{9887975}
X.~Wang, D.~Zhang, H.-Z. Tan, and D.-J. Lee, ``A self-fusion network based on
  contrastive learning for group emotion recognition,'' \emph{IEEE Transactions
  on Computational Social Systems}, pp. 1--12, 2022.

\bibitem{liu2019roberta}
Y.~Liu, M.~Ott, N.~Goyal, J.~Du, M.~Joshi, D.~Chen, O.~Levy, M.~Lewis,
  L.~Zettlemoyer, and V.~Stoyanov, ``Roberta: A robustly optimized bert
  pretraining approach,'' \emph{arXiv preprint arXiv:1907.11692}, 2019.

\bibitem{huang2017densely}
G.~Huang, Z.~Liu, L.~Van Der~Maaten, and K.~Q. Weinberger, ``Densely connected
  convolutional networks,'' in \emph{2017 IEEE Conference on Computer Vision
  and Pattern Recognition (CVPR)}.\hskip 1em plus 0.5em minus 0.4em\relax IEEE,
  2017, pp. 2261--2269.

\bibitem{NEURIPS2018_6832a7b2}
Y.~Jia, Y.~Zhang, R.~Weiss, Q.~Wang, J.~Shen, F.~Ren, z.~Chen, P.~Nguyen,
  R.~Pang, I.~Lopez~Moreno, and Y.~Wu, ``Transfer learning from speaker
  verification to multispeaker text-to-speech synthesis,'' in \emph{Advances in
  Neural Information Processing Systems}, vol.~31.\hskip 1em plus 0.5em minus
  0.4em\relax MIT, 2018.

\bibitem{oria-etal-2019-meld}
S.~Poria, D.~Hazarika, N.~Majumder, G.~Naik, E.~Cambria, and R.~Mihalcea,
  ``{MELD}: A multimodal multi-party dataset for emotion recognition in
  conversations,'' pp. 527--536, 2019.

\bibitem{busso2008iemocap}
C.~Busso, M.~Bulut, C.-C. Lee, A.~Kazemzadeh, E.~Mower, S.~Kim, J.~N. Chang,
  S.~Lee, and S.~S. Narayanan, ``Iemocap: Interactive emotional dyadic motion
  capture database,'' \emph{Language resources and evaluation}, vol.~42, no.~4,
  pp. 335--359, 2008.

\bibitem{kim-2014-convolutional}
Y.~Kim, ``Convolutional neural networks for sentence classification,'' in
  \emph{Proceedings of the 2014 Conference on Empirical Methods in Natural
  Language Processing ({EMNLP})}.\hskip 1em plus 0.5em minus 0.4em\relax ACL,
  2014, pp. 1746--1751.

\bibitem{ghosal-etal-2019-dialoguegcn}
D.~Ghosal, N.~Majumder, S.~Poria, N.~Chhaya, and A.~Gelbukh, ``{D}ialogue{GCN}:
  A graph convolutional neural network for emotion recognition in
  conversation,'' in \emph{Proceedings of the 2019 Conference on Empirical
  Methods in Natural Language Processing and the 9th International Joint
  Conference on Natural Language Processing (EMNLP-IJCNLP)}.\hskip 1em plus
  0.5em minus 0.4em\relax ACL, 2019, pp. 154--164.

\bibitem{9316758}
Z.~Lian, B.~Liu, and J.~Tao, ``Ctnet: Conversational transformer network for
  emotion recognition,'' \emph{IEEE/ACM Transactions on Audio, Speech, and
  Language Processing}, vol.~29, pp. 985--1000, 2021.

\end{thebibliography}

\vfill

\end{document}